\documentclass{article}

\usepackage[final, nonatbib]{neurips_2020}
\usepackage[utf8]{inputenc} 
\usepackage[T1]{fontenc}    
\usepackage{hyperref}       
\usepackage{url}            
\usepackage{booktabs}       
\usepackage{amsfonts}       
\usepackage{nicefrac}       
\usepackage{microtype}      
\usepackage{graphicx}
\usepackage{caption}
\usepackage{subcaption}

\usepackage{float}
\usepackage{amsmath,amsfonts,amssymb}
\newcommand{\E}{\mathbb{E}}       
\newcommand{\R}{\mathbb{R}}       
\newcommand{\vct}[1]{\mathbf{#1}} 
\newcommand{\mat}[1]{\mathbf{#1}} 

\title{Improved Training of Sparse Coding Variational Autoencoder via Weight Normalization}

\author{%
  Linxing Preston Jiang, Luciano de la Iglesia \\
  Paul G. Allen School of Computer Science \& Engineering \\
  University of Washington\\
  \texttt{\{prestonj,lucianod\}@cs.washington.edu} \\
}

\begin{document}

\maketitle

\begin{abstract}
    Learning a generative model of visual information with sparse and compositional features has been a challenge for both theoretical neuroscience and machine learning communities. Sparse coding models have achieved great success in explaining the receptive fields of mammalian primary visual cortex with sparsely activated latent representation. In this paper, we focus on a recently proposed model, sparse coding variational autoencoder (SVAE) (Barello et al., 2018), and show that the end-to-end training scheme of SVAE leads to a large group of decoding filters not fully optimized with noise-like receptive fields. We propose a few heuristics to improve the training of SVAE and show that a unit $L_2$ norm constraint on the decoder is critical to produce sparse coding filters. Such normalization can be considered as local lateral inhibition in the cortex. We verify this claim empirically on both natural image patches and MNIST dataset and show that projection of the filters onto unit norm drastically increases the number of active filters. Our results highlight the importance weight normalization for learning sparse representation from data and suggest a new way of reducing the number of inactive latent components in VAE learning.
\end{abstract}

\section{Introduction}

One key challenge in theoretical neuroscience is to understand the computation carried through the visual pathways. Hubel and Wiesel first showed that neurons in mammalian primary visual cortex (V1) have spatially localized and orientation-selective receptive fields \cite{hubel_receptive_1959} that interestingly resemble edge detectors or ``parts'' of objects. This connection between visual neurons' neurophysiological properties and the statistics of the environment was successfully explore by Olshausen \& Field \cite{olshausen_emergence_1996}. In their model, they proposed that the primary visual cortex is learning a generative model of the visual world with sparsely activated neural activities. Remarkably, such a model produces edge detector style Gabor-like spatial filters that are similar to V1 receptive fields. This is known as the ``sparse coding'' model of V1. 

Sparse coding (or sparse dictionary learning) has been extensively studied in the machine learning community as an unsupervised generative model of images \cite{Lee2006EfficientSC, Gregor2010LearningFA, Chen2018TheSM}. It has also been shown to be robust to adversial attacks \cite{Paiton2020SelectivityAR, Sulam2020AdversarialRO}. More recently, Barello et al. \cite{Barello2018SparseCodingVA} proposed a sparse coding variational autoencoder (SVAE) model that combines a variational autoencoder (VAE) \cite{Kingma2014AutoEncodingVB} with a sparse coding decoder for learning sparse structure in data. Compared to the original sparse coding model, SVAE shows better reconstruction performances and allows stochastic latent representations, which is more neurally plausible than a deterministic maximum a posteriori (MAP) estimate in traditional sparse coding.

Our project focused on improving the quality of the learned decoder in SVAE. We show empirically that the formulation and training of SVAE leads to a large portion of unoptimized filters in the decoder, commonly known as the ``over-pruning'' problem in VAEs \cite{Bowman2016GeneratingSF, Kingma2017ImprovedVI, Yeung2017TacklingOI}. We propose three heuristics to improve the training of SVAE: First, we weighed the Kullback–Leibler (KL) divergence term in the loss function by $\beta$, a technique proposed by the $\beta$-VAE model \cite{Higgins2017betaVAELB}. Using a $\beta$ term less than 1 smooths the effect of the sparsity prior, causing large gradients on only a few filters. Second, we used a more expressive encoder architecture with ResNet blocks \cite{He2016DeepRL} to replace the linear filters in the original SVAE for better posterior approximation. Most importantly, we applied
the same projected gradient descent step in the original sparse coding to constrain the decoder filters to have unit length. This constraint drastically improves the number of filters that are optimized and have similar quality receptive fields to the sparse coding model. We validate our claims by comparing the performance of the original SVAE and our training procedure on natural images patches. SVAE trained with our approach shows similar reconstruction error to the original training and produced qualitatively better filters that resemble parts of images.
We further show that the unit-length constraint of the filters (which can be viewed as lateral inhibition in the cortex) is critical for the formation of Gabor-like filters on both natural images and MNIST dataset.

\section{Background}
We introduce the formulation of sparse coding and sparse coding variational autoencoders (SVAEs) in this section. 
\subsection{Sparse coding}

The sparse coding model minimizes the following energy function
\begin{align*}
    \min_{\mat{U}, \vct{z}} E &= \Vert \vct{x} - \mat{Uz} \Vert_2^2 + \lambda \Vert\vct{z}\Vert_1  \\
        &\text{s.t.}  \Vert\mat{U}_i\Vert_2 \leq 1 && \forall i = 1, 2, \dots, N
\end{align*} 
where $\vct{x}\in\R^D$ denotes the input, $\mat{U}\in\R^{D \times N}$ represents the receptive fields (RFs) or filters of the model, and $\vct{z} \in \R^N$ represents the neural activation (latent variables). The $L_1$ penalty on $\vct{z}$ is a relaxation of the $L_0$ penalty that promotes sparsity in $\vct{z}$ (only a small subset of the components is nonzero). $\lambda$ is a scalar that controls the degree of the sparsity penalty. In addition, each filter (column of $\vct{U}$) is constrained to have unit $L_2$ norm to prevent a few filters with large weights from dominating image reconstruction. We will show in later sections and results that this is a key constraint to promote filter quality and increase the number of active latent variables.

\paragraph{Inference} A common practice is to use proximal gradient descent rather than the vanilla gradient descent for faster convergence of the latent code. We use the iterative shrinkage threshold algorithm (ISTA) \cite{Boyd2006ConvexO} which takes a shrinkage step after a gradient update. The gradient update is defined as:
\begin{align*}
    \frac{\partial E}{\partial\vct{z}} = -2\mat{U}^{\intercal}(\vct{x} - \mat{U}\vct{z})
\end{align*}
and the shrinkage update is defined as
\begin{align*} 
    \vct{z}' &=\operatorname{Shrinkage}_{\lambda}(\vct{z})  \\
    &=\operatorname{sign}(\vct{z})\max(|\vct{z}|-\lambda, 0)
\end{align*}
We consider $\vct{z}$ as converged if the change of its $L_2$ norm before and after one update is less than 1\%. 

\paragraph{Learning} After $\vct{r}$ converges for the current input $\vct{x}$, we update $\mat{U}$ using projected gradient descent. The update rule is defined as 
\begin{align*}
    \frac{\partial E}{\partial\mat{U}} = -2(\vct{x} - \mat{U}\vct{z})\vct{z}^{\intercal}
\end{align*}
where $\eta$ is the learning rate. After a gradient update, we project each column of $\mat{U}$ to unit norm following the constraint. 

\subsection{VAE and SVAE}

The inference step of sparse coding can be seen as performing MAP estimate through gradient updates. Variational autoencoders (VAEs), on the other hand, perform inference using a feedforward mapping from the observation to the latent posterior using shared parameters across observations. Such mapping is then learned through minimizing the Kullback–Leibler (KL) divergence between the approximating distribution and the true posterior
\begin{align*}
    D_{KL}(q(\vct{z} | \vct{x}) \Vert p(\vct{z} | \vct{x}))
\end{align*}
Since we cannot tractably compute the true posterior, VAEs choose to optimize the evidence lower bound (ELBO) $\mathcal{L}$ defined as 
\begin{align*}
    D_{KL}(q(\vct{z} | \vct{x}) \Vert p(\vct{z} | \vct{x})) = \log p_\theta(\vct{x}) - \mathcal{L}
\end{align*}
\begin{align*}
    \mathcal{L} &= \E_{\vct{z} \sim q(\cdot | \vct{x})}\left[\log p_{\theta}(\mathbf{x}, \mathbf{z})-\log q(\mathbf{z} | \mathbf{x})\right] \\
        &= \E_{\vct{z} \sim q(\cdot | \vct{x})}\left[\log p_{\theta}(\mathbf{x} | \mathbf{z})\right]-D_{K L}\left(q(\mathbf{z} | \mathbf{x}) \| p_{\theta}(\mathbf{z})\right)
\end{align*}

In the original VAE formulation \cite{Kingma2014AutoEncodingVB}, the proposal distribution $q(\vct{z} | \vct{x})$, the latent prior $p_\theta(\vct{z})$, and the likelihood term $p_\theta(\vct{x} | \vct{z})$ are all chosen to be Gaussian. The dimension of the latent code $\vct{z}$ is typically much smaller than the data dimension, forcing the model to learn low-dimensional structures that generate the true data distribution.

\paragraph{SVAE} In the work of Barello et al. \cite{Barello2018SparseCodingVA}, the authors proposed three modifications to the original VAE: (1) Make the dimension of the latent variables overcomplete (larger than the input dimension); (2) Use a sparsity-inducing prior (e.g. $p_\theta(\vct{z}) \sim \text{Laplace}(0, 1)$) instead of the Gaussian prior; (3) Parameterize the decoder with a single linear layer rather than a deep neural network. In SVAE, the encoder replaces the iterative inference (ISTA), generating a full posterior $q(\vct{z} | \vct{x})$ instead of a single MAP estimation. The decoder behaves like the sparse coding filters $\mat{U}$ by taking a sampled $\vct{z} \sim q(\vct{z} | \vct{x})$ and reconstructing the input $\vct{\hat{x}} = \mat{U}\vct{z}$. The proposal distribution and the likelihood term remain Gaussian, and the encoder is parameterized with two linear layers followed by a ReLU nonlinearity, and two separate linear layers that generate the mean and the log variance of the proposal distribution, respectively. 

However, in practice, we found that this SVAE formulation leads to a large number of noise filters learned in the decoder. Figure \ref{receptive_fields} shows the decoder filters learned by sparse coding (left) and SVAE (middle). Only a small subset of filters in SVAE decoder resemble the oriented bandpass Gabor filters like the RFs in V1, while most of the filters are not optimized to represent sparse structure in natural image data. In the next section, we present a few heuristics to improve the number of active filters learned in SVAE (Figure \ref{receptive_fields}, right). 

\section{Improving SVAE training}

To tackle the issue of under-optimized filters in the decoder of SVAE, we propose the following heuristics to improve the training of SVAE. 

\paragraph{Balancing between reconstruction and KL divergence} Following the idea proposed in $\beta$-VAE \cite{Higgins2017betaVAELB}, we weigh the KL divergence term in ELBO to be less than 1 in order to smooth the effect of the sparse prior placing heavy gradients on only few filters. The new ELBO term then becomes 
\begin{align*}
    \mathcal{L} = \E_{\vct{z} \sim q(\cdot | \vct{x})}\left[\log p_{\theta}(\mathbf{x} | \mathbf{z})\right]-\beta D_{K L}\left(q(\mathbf{z} | \mathbf{x}) \| p_{\theta}(\mathbf{z})\right)
\end{align*}

\paragraph{More expressive encoder} To improve the quality of the approximating posterior, we replace the linear layers in the original SVAE with ResNet blocks \cite{He2016DeepRL}. 

\paragraph{SVAE decoder with unit norm constraint} Most importantly, we noticed that in the end-to-end training of SVAE, the filters of the decoder no longer have the constraint of having unit $L_2$ norm. We suspect that the over-pruning issue of VAE training \cite{Yeung2017TacklingOI} and the sparsity-inducing prior together exacerbate the imbalance of training, which causes only a few filters to have dominating gradients, leaving most of the filters not fully optimized. Therefore, we propose to use the same projected gradient descent step in the original sparse coding model on SVAE to ensure each decoder filter is constrained to unit $L_2$ norm. We show that this drastically improves the number of optimized filters (see Results). 

\begin{figure}[t]
  \includegraphics[width=\textwidth]{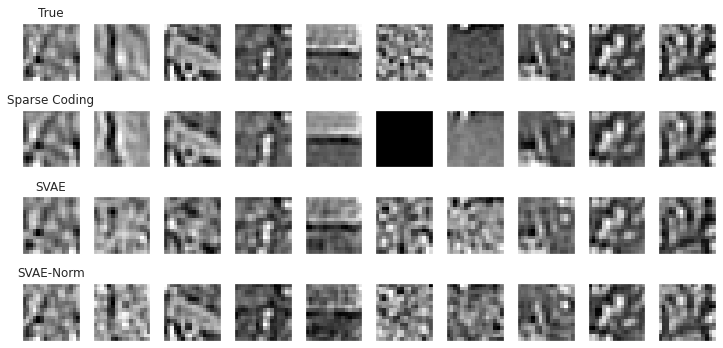}
  \caption{Reconstructions of 16x16 natural image patches in the held-out test set using the different models.}
  \label{reconstructions}
\end{figure}
\section{Results}
Here, we present the performance comparison among the traditional sparse coding model, SVAE, and SVAE with our training heuristics. To further investigate the effect of decoder weight normalization, we also trained two SVAEs both with the first two proposed heuristics ($\beta$ weighting, ResNet block encoders), but one with weight normalization ("SVAE-Norm") and the other one without. We experimented with natural image patches and MNIST handwritten digits. The natural image patches\footnote{\url{http://www.rctn.org/bruno/sparsenet/}} are spatially whitened with a low pass filter $R(f) = fe^{-(f/f_0)^4}$, $f_0 = 200$ cycles/image. We used $z_i \sim \text{Laplace}(0, 0.1) \text{ }\forall i = 1\dots N$ as the factorized sparsity-inducing prior for all experiments. 
\label{results}
\subsection{Reconstruction}
We first examined the performance of the models on reconstruction. Figure \ref{reconstructions}
shows 10 reconstructed patches, while Table \ref{mse} shows the pixelwise mean
squared error (MSE) of each model's reconstructions on the entire test set.
Our normalized SVAE model (SVAE-Norm) achieves the lowest MSE, followed by 
SVAE. The traditional sparse coding model performs worse than both.
\begin{table}[h]
  \centering
  \begin{tabular}{ c|c|c }
    
    Model & MSE & STD (Monte Carlo Samples) \\
    \hline
    Sparse Coding & 0.0150 & N/A \\
    SVAE & 0.00875 & 1.631e-5 \\
    SVAE-Norm & 0.00769 & 2.329e-5 \\
    
  \end{tabular}
  \caption{The mean squared error and standard deviation for reconstructions of natural images in the held-out test set over 50 trials.}
  \label{mse}
\end{table}
\subsection{Neural representation and activity}
\begin{figure}
  \includegraphics[width=\textwidth]{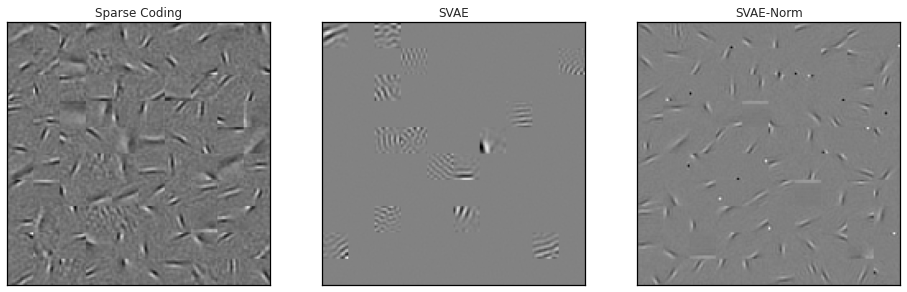}
  \caption{Receptive fields for 100 neurons in the representation layer of each of the models.}
  \label{receptive_fields}
\end{figure}

Next, we examined the receptive fields of the neurons in model's decoder, shown in Figure~\ref{receptive_fields}. In the SVAE formulation, these filters should behave similarly to the traditional sparse coding filters. SVAE-Norm exhibits the clearest Gabors, followed by the sparse coding model. SVAE has some Gabor-like structures, but most of its neurons have either PCA-like receptive fields \cite{olshausen_emergence_1996} (e.g. Figure~\ref{receptive_fields} middle, first row, middle column) or are unoptimized (gray filters). 

\begin{figure}[H]
  \includegraphics[width=\textwidth]{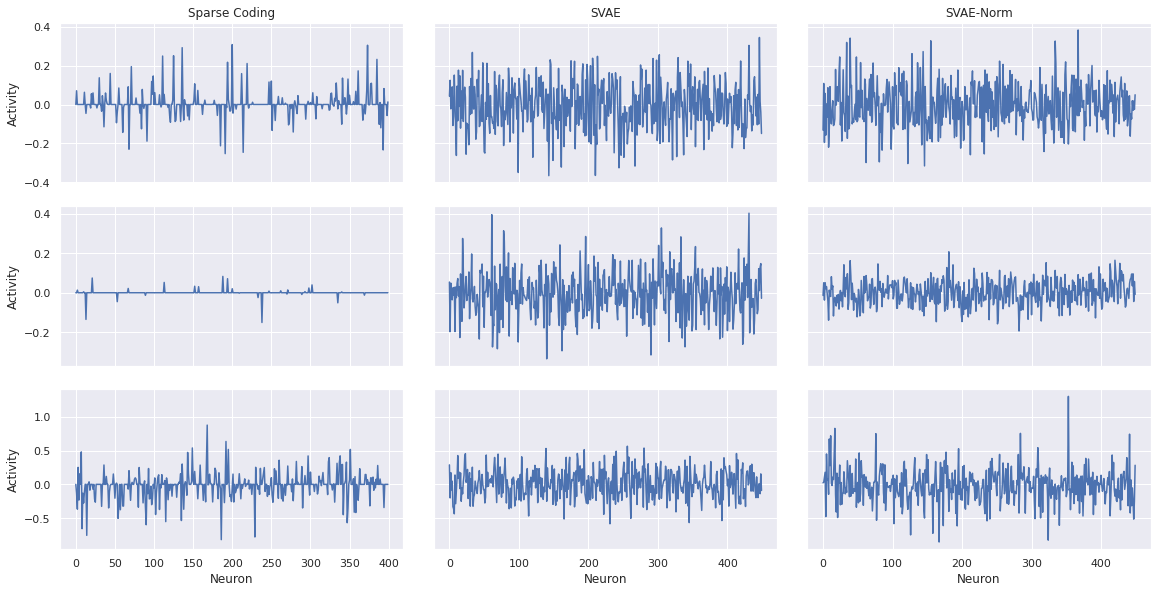}
  \caption{Activity of all the neurons in the representation layer of our networks for
  3 different natural image patch inputs shown in each row.}
  \label{activations}
\end{figure}

When presented with a natural image patch, only a few neurons have a large activation (Figure~\ref{activations}, each row shows a different input patch). This is consistent with biological and computational findings in the sparse-coding domain. However, the sparse coding model has a clearer sparse activation across the 3 images (left column), while the VAEs visually appear to be noisier (middle and right column). This is expected due to the different inference procedures (ISTA vs. amortized inference + sampling), although it is interesting to see that strictly sparse code (true zero activations) is not required to produce ``parts of images'' structures in the filters (in SVAEs).

\begin{figure}[t]
  \includegraphics[width=\textwidth]{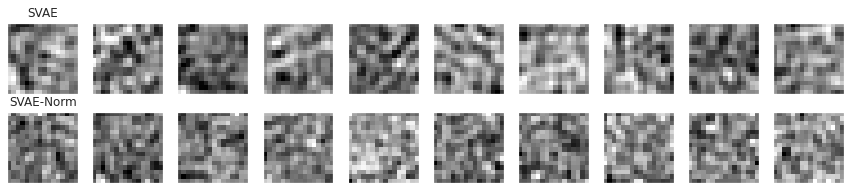}
  \caption{Images generated by random sampling from the prior distribution of our
  VAE models.}
  \label{samples}
\end{figure}

To evaluate the conditional distribution $p_\theta(\vct{x} | \vct{z})$ learned by the decoder, we generated images by sampling from the prior Laplace distribution (Figure~\ref{samples}).
Although these images are noisy, they look similar to the natural image patches of Figure
\ref{reconstructions}, suggesting the SVAEs learned a meaningful hidden representation that generates input data.

\begin{figure}[h]
  \begin{subfigure}[t]{0.55\textwidth}
    \includegraphics[width=\textwidth]{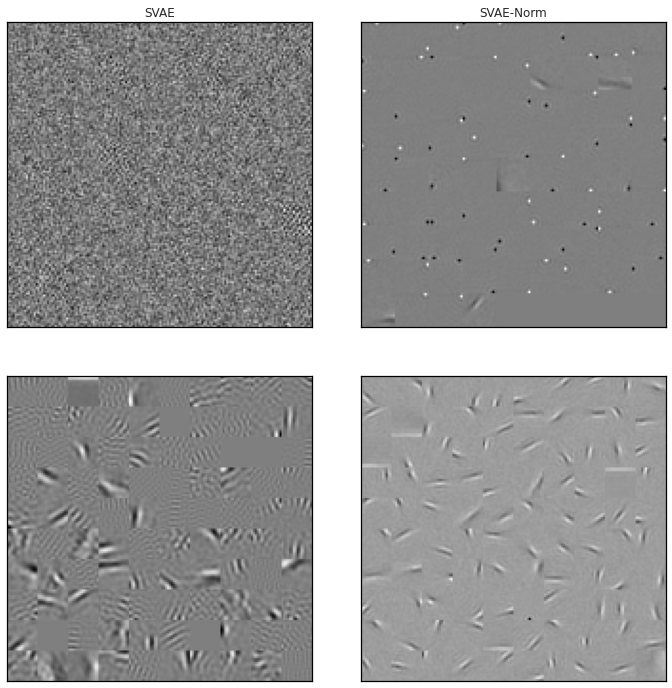}
    \caption{The receptive fields of neurons in the SVAE-Norm model (right) and the original SVAE (left) are shown. The top shows those which did not achieve activity greater than 0.5 for any image in the test set, and the bottom shows the rest. This separates the filters into noise (top) and Gabor (bottom) groups, with up to 100 randomly selected neurons shown. The SVAE has 154 Gabor and 296 noise neurons, while the SVAE-Norm has 374 Gabor and 76 noise neurons.}
  \end{subfigure}
  \hfill
  \begin{subfigure}[t]{0.43\textwidth}
    \includegraphics[width=\textwidth]{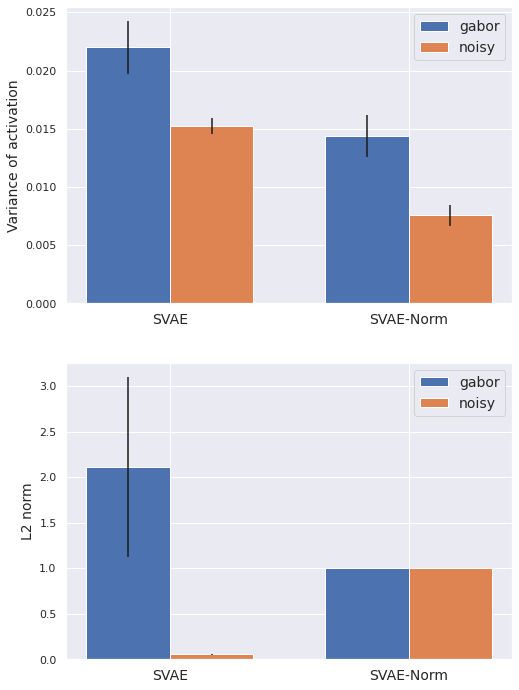}
    \caption{The top plot shows variance of activation across the test set for the 2 groups of neurons in
    both models. The bottom plot shows the $L_2$ norm for both groups and models. Note in the SVAE-Norm model
    the vectors were normalized to 1.}
  \end{subfigure}
  \caption{Effect of weight normalization on SVAE decoder. (a) Example of Gabor-like vs. noisy filters; (b) Noisy filters show smaller variances on test set with extremely small norm length compared to the Gabor-like filters.}
  \label{grouped_analysis}
\end{figure}

\subsection{Noise filters}
We observed that many neurons are under-optimized and learned white-noise-like filters rather than Gabor-like filters in the SVAE model (Figure~\ref{grouped_analysis}(a), top left). This suggests that a majority of the filters were not learning meaningful latent structure of the data. We were able to isolate these neurons in the SVAE model by thresholding based on the standard deviation of the latent code activation on the test set. We used 0.5 as the threshold for both SVAE and SVAE-Norm model, and found that 296 filters out of 450 in SVAE are below the threshold,
We visualized the variance distribution of these filters in Figure~\ref{grouped_analysis}(b) top panel. In Figure~\ref{grouped_analysis}(b), we show that these "noise" filters have extremely small $L_2$ norms, compared to the filters that show clear Gabor-like structure (Figure~\ref{grouped_analysis}(a), bottom left). However, with weight normalization applied to SVAE during training, only 76 filters out of 450 are below the threshold, and these filters are mainly highly sparse filters that encode single pixels of the images (Figure~\ref{grouped_analysis}(a), top right). The majority of the filters in SVAE-Norm model now show Gabor-like structure similar to the traditional sparse coding models and V1 RFs ((Figure~\ref{grouped_analysis}(a), bottom right). 

\begin{figure}
    \centering
    \includegraphics[width=\textwidth]{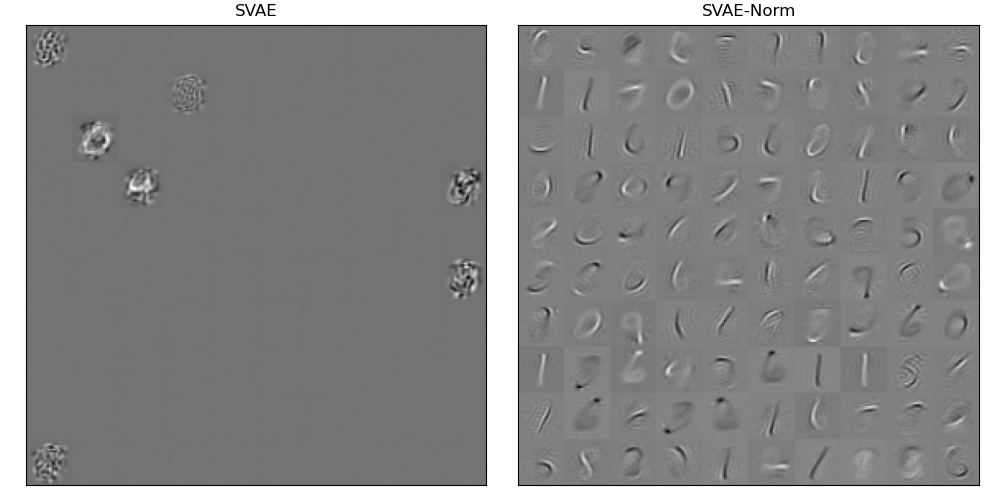}
    \caption{100 randomly sampled receptive fields trained from MNIST data. Left: SVAE; Right: SVAE with weight normalization}
    \label{fig:mnist}
\end{figure}

To validate the effect of weight normalization in learning quality filters, we ran the same two SVAE models on the MNIST datasets, with or without weight normalization. Figure~\ref{fig:mnist} shows 100 randomly sampled decoder filters from the two models. The original SVAE model only shows 7 filters with large norms but noisy structures, while all filters in the SVAE-Norm model show stroke-like structures that resemble parts of MNIST digits. For a visualization of all of the filters learned by the two models on the two datasets, see Appendix figures.

\section{Discussion}
In this paper, we showed that a weight normalization scheme for training sparse coding variational autoencoders is critical for decoder optimization. To gain some insights into the efficacy of the approach, the projected gradient descent on the decoder can be thought of as a special case of the weight normalization proposed by Salimans et al. \cite{Salimans2016WeightNA}. The weight normalization trick reparameterizes neural network weights $\vct{w}$ as 
\begin{align*}
    \vct{w} = \frac{g}{\Vert\vct{v}\Vert_2}\vct{v}
\end{align*}
where $g$ defines the length of the unit vector $\frac{\vct{v}}{\Vert\vct{v}\Vert_2}$. The SVAE decoder can be thought of as reparameterized weights with $g = 1$. Weight normalization has been shown to accelerate model training and encourage disentangled representation learning. We expect having a unit norm constraint on the SVAE decoder to have similar effects. Future work could focus on verifying the effect of normalization on hierarchical latent variables models, e.g. extending hierarchical sparse coding \cite{Zeiler2010DeconvolutionalN, Zeiler2011AdaptiveDN} in a VAE setting. 

\section{Acknowledgements}
This paper was written as a final project for a course on generative models at the University of Washington. We thank the course instructor John Thickstun for his feedback.

\bibliographystyle{unsrt}
\bibliography{ref}

\begin{thebibliography}{10}

\bibitem{hubel_receptive_1959}
D.~H. Hubel and T.~N. Wiesel.
\newblock Receptive fields of single neurones in the cat's striate cortex.
\newblock {\em The Journal of Physiology}, 148(3):574--591, October 1959.

\bibitem{olshausen_emergence_1996}
Bruno~A. Olshausen and David~J. Field.
\newblock Emergence of simple-cell receptive field properties by learning a
  sparse code for natural images.
\newblock {\em Nature}, 381(6583):607--609, June 1996.
\newblock Number: 6583 Publisher: Nature Publishing Group.

\bibitem{Lee2006EfficientSC}
H.~Lee, A.~Battle, R.~Raina, and A.~Ng.
\newblock Efficient sparse coding algorithms.
\newblock In {\em NIPS}, 2006.

\bibitem{Gregor2010LearningFA}
K.~Gregor and Y.~LeCun.
\newblock Learning fast approximations of sparse coding.
\newblock In {\em ICML}, 2010.

\bibitem{Chen2018TheSM}
Y.~Chen, Dylan~M. Paiton, and B.~Olshausen.
\newblock The sparse manifold transform.
\newblock {\em ArXiv}, abs/1806.08887, 2018.

\bibitem{Paiton2020SelectivityAR}
Dylan~M. Paiton, Charles~G. Frye, Sheng~Y. Lundquist, Joel~D Bowen, R.~Zarcone,
  and B.~Olshausen.
\newblock Selectivity and robustness of sparse coding networks.
\newblock {\em Journal of Vision}, 20, 2020.

\bibitem{Sulam2020AdversarialRO}
Jeremias Sulam, Ramchandran Muthumukar, and R.~Arora.
\newblock Adversarial robustness of supervised sparse coding.
\newblock {\em ArXiv}, abs/2010.12088, 2020.

\bibitem{Barello2018SparseCodingVA}
G.~Barello, A.~Charles, and Jonathan~W. Pillow.
\newblock Sparse-coding variational auto-encoders.
\newblock {\em bioRxiv}, 2018.

\bibitem{Kingma2014AutoEncodingVB}
Diederik~P. Kingma and M.~Welling.
\newblock Auto-encoding variational bayes.
\newblock {\em CoRR}, abs/1312.6114, 2014.

\bibitem{Bowman2016GeneratingSF}
Samuel~R. Bowman, L.~Vilnis, Oriol Vinyals, Andrew~M. Dai, R.~J{\'o}zefowicz,
  and S.~Bengio.
\newblock Generating sentences from a continuous space.
\newblock {\em ArXiv}, abs/1511.06349, 2016.

\bibitem{Kingma2017ImprovedVI}
Diederik~P. Kingma, Tim Salimans, and M.~Welling.
\newblock Improved variational inference with inverse autoregressive flow.
\newblock {\em ArXiv}, abs/1606.04934, 2017.

\bibitem{Yeung2017TacklingOI}
Serena Yeung, A.~Kannan, Yann Dauphin, and Li~Fei-Fei.
\newblock Tackling over-pruning in variational autoencoders.
\newblock {\em ArXiv}, abs/1706.03643, 2017.

\bibitem{Higgins2017betaVAELB}
I.~Higgins, Lo{\"i}c Matthey, A.~Pal, C.~Burgess, Xavier Glorot, M.~Botvinick,
  S.~Mohamed, and Alexander Lerchner.
\newblock beta-vae: Learning basic visual concepts with a constrained
  variational framework.
\newblock In {\em ICLR}, 2017.

\bibitem{He2016DeepRL}
Kaiming He, X.~Zhang, Shaoqing Ren, and Jian Sun.
\newblock Deep residual learning for image recognition.
\newblock {\em 2016 IEEE Conference on Computer Vision and Pattern Recognition
  (CVPR)}, pages 770--778, 2016.

\bibitem{Boyd2006ConvexO}
Stephen~P. Boyd and L.~Vandenberghe.
\newblock Convex optimization.
\newblock {\em IEEE Transactions on Automatic Control}, 51:1859--1859, 2006.

\bibitem{Salimans2016WeightNA}
Tim Salimans and Diederik~P. Kingma.
\newblock Weight normalization: A simple reparameterization to accelerate
  training of deep neural networks.
\newblock In {\em NIPS}, 2016.

\bibitem{Zeiler2010DeconvolutionalN}
Matthew~D. Zeiler, Dilip Krishnan, Graham~W. Taylor, and R.~Fergus.
\newblock Deconvolutional networks.
\newblock {\em 2010 IEEE Computer Society Conference on Computer Vision and
  Pattern Recognition}, pages 2528--2535, 2010.

\bibitem{Zeiler2011AdaptiveDN}
Matthew~D. Zeiler, Graham~W. Taylor, and R.~Fergus.
\newblock Adaptive deconvolutional networks for mid and high level feature
  learning.
\newblock {\em 2011 International Conference on Computer Vision}, pages
  2018--2025, 2011.

\end{thebibliography}

\newpage
\section*{Appendix}
\begin{figure}[H]
    \centering
    \includegraphics[width=1\textwidth]{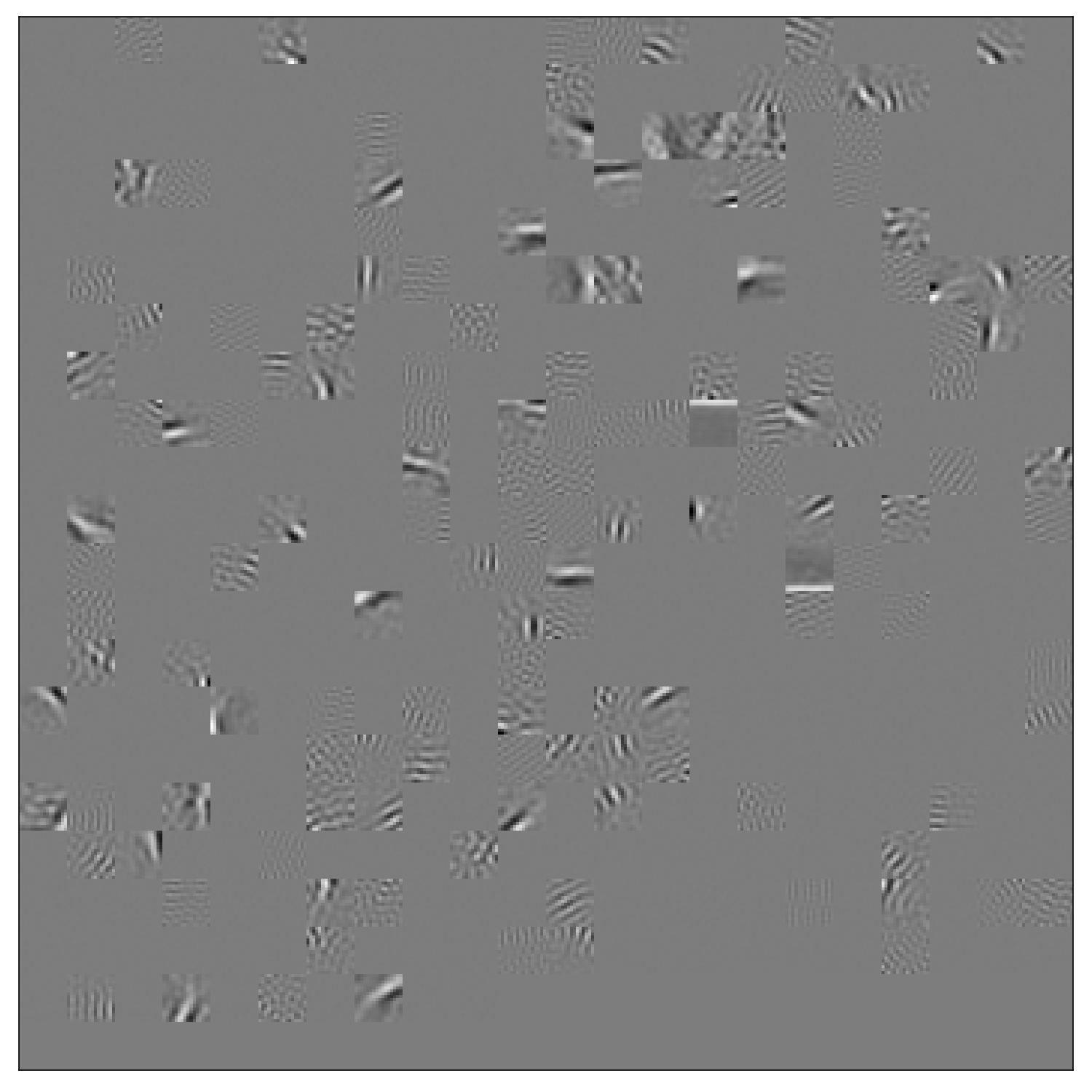}
    \caption{All SVAE decoder filters learned on natural image patches}
    \label{fig:rf_svae_full}
\end{figure}
\newpage
\begin{figure}[H]
    \centering
    \includegraphics[width=1\textwidth]{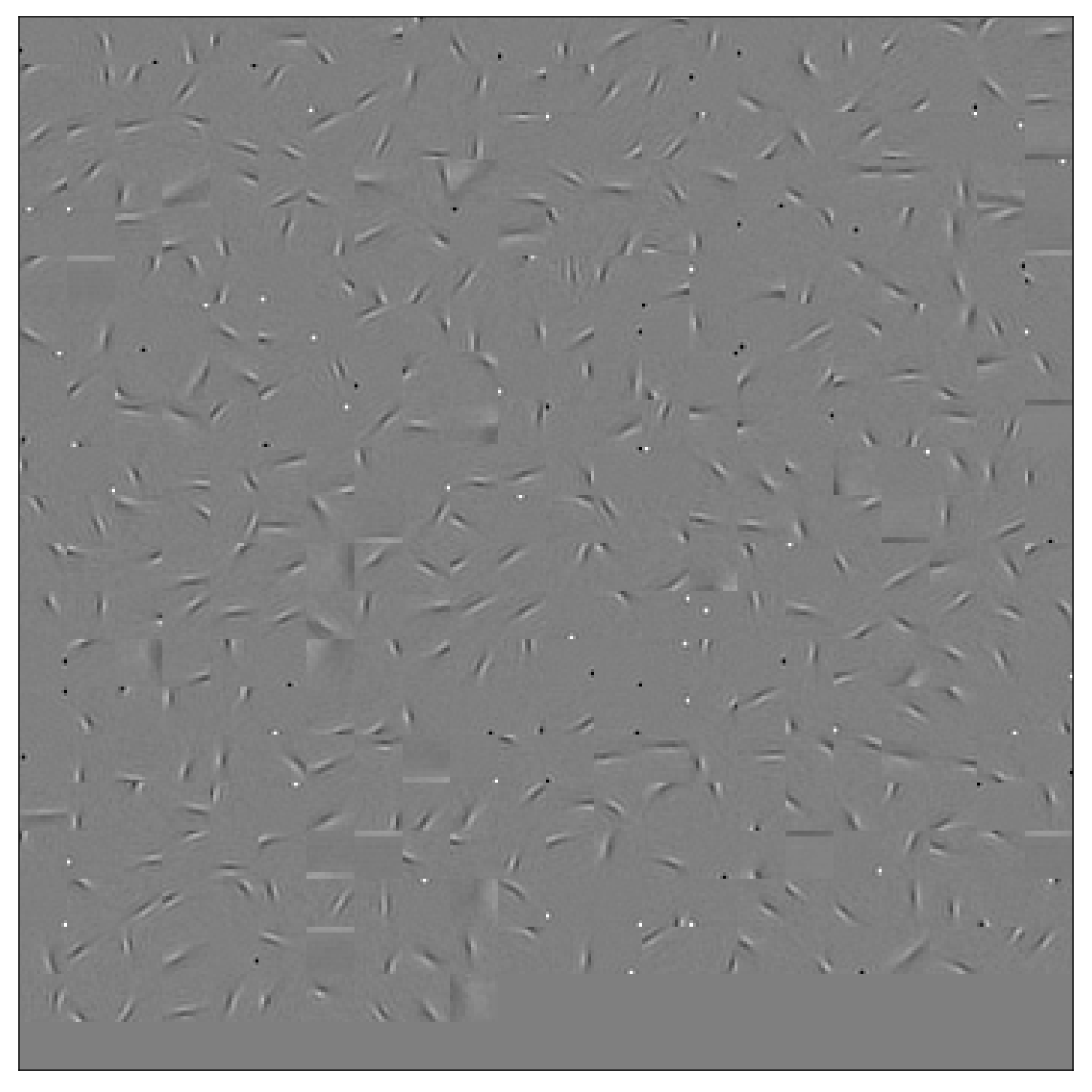}
    \caption{All SVAE decoder filters with weight normalization learned on natural image patches}
    \label{fig:rf_svae_norm_full}
\end{figure}
\begin{figure}[H]
    \centering
    \includegraphics[width=1.2\textwidth]{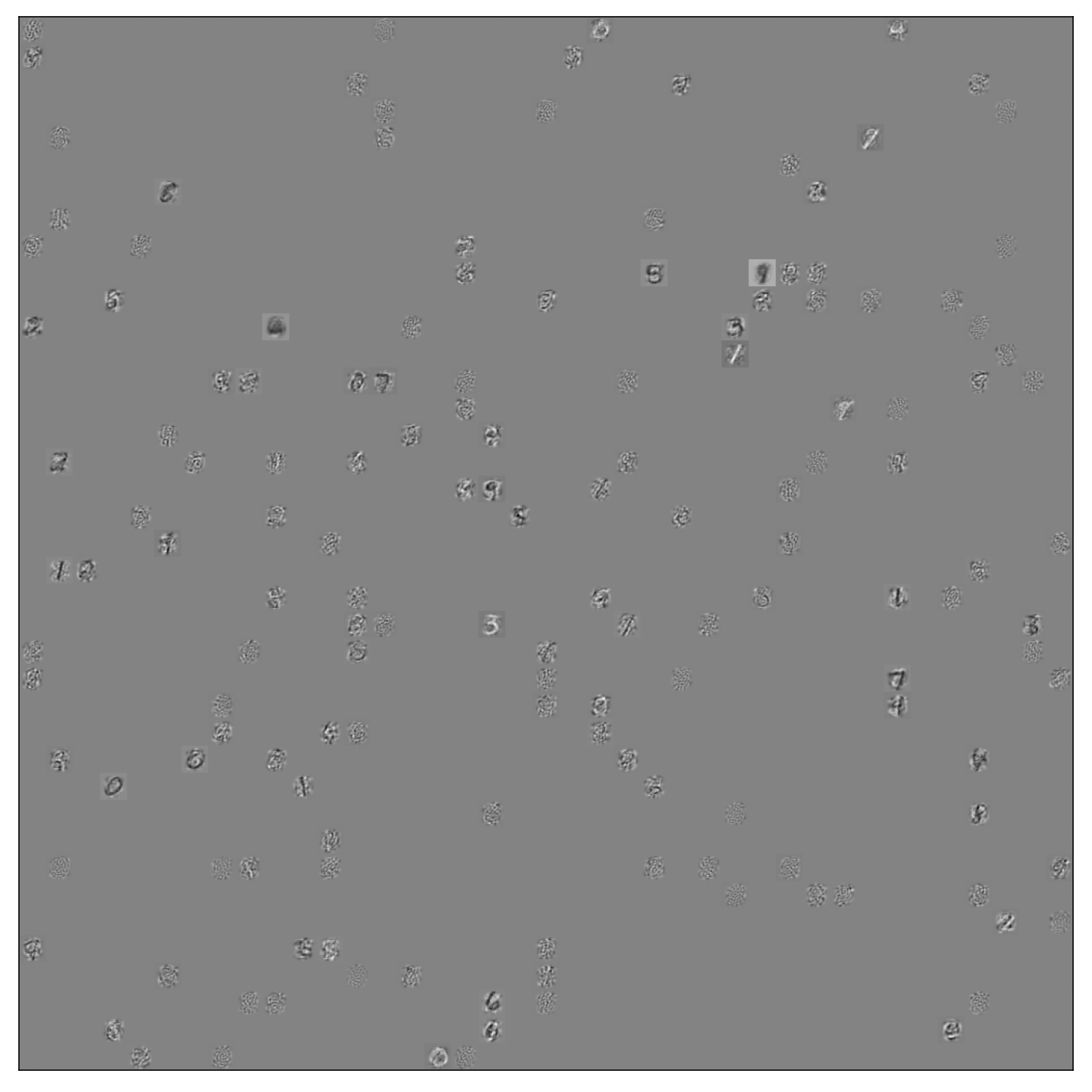}
    \caption{All SVAE decoder filters learned on MNIST}
    \label{fig:rf_svae_mnist_full}
\end{figure}
\newpage
\begin{figure}[H]
    \centering
    \includegraphics[width=1.2\textwidth]{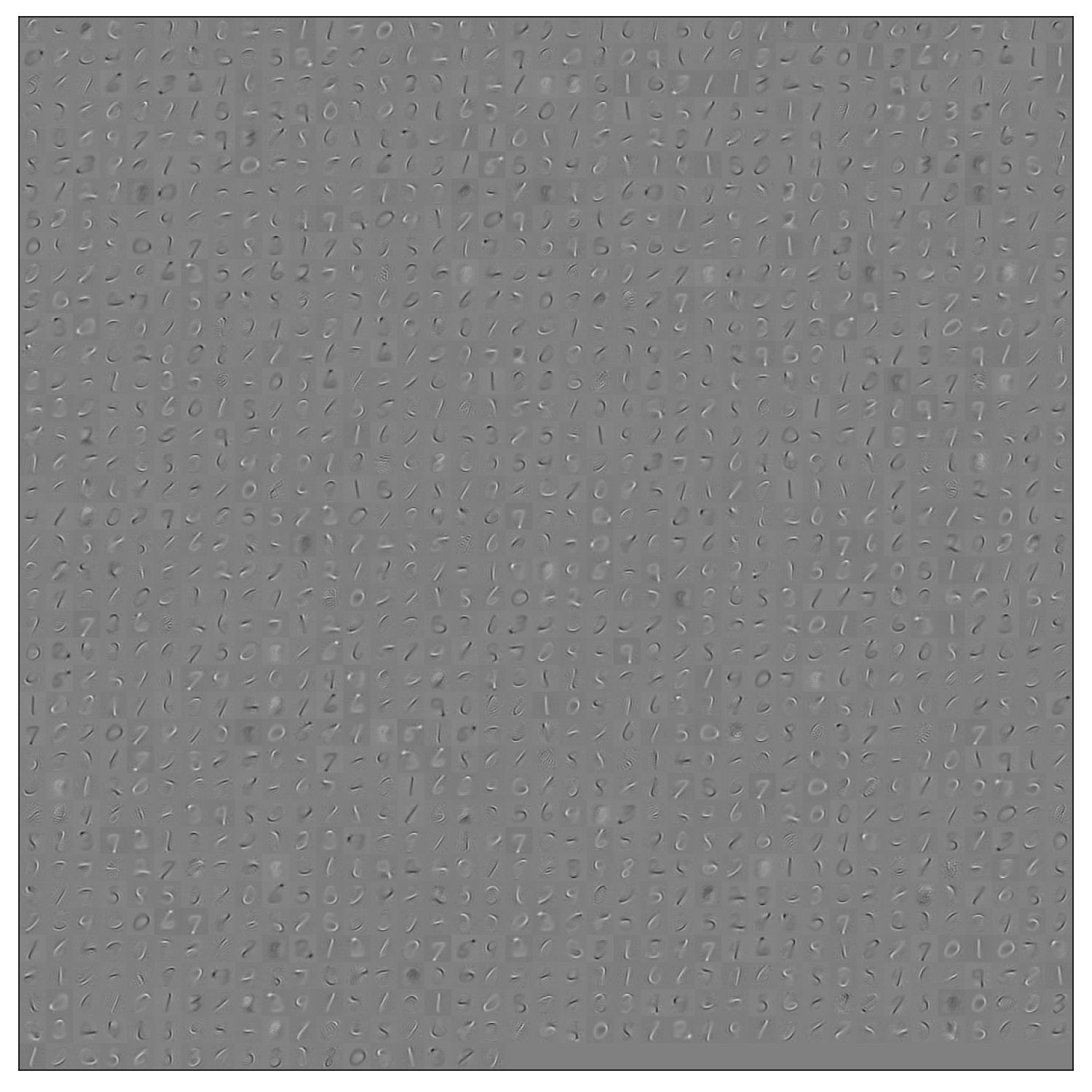}
    \caption{All SVAE decoder filters with weight normalization learned on MNIST}
    \label{fig:rf_svae_norm_mnist_full}
\end{figure}
\end{document}